\documentclass{article}

\usepackage{microtype}
\usepackage{graphicx}
\usepackage{subfigure}
\usepackage{booktabs} 

\usepackage{hyperref}

\usepackage[accepted]{icml2025}

\usepackage{amsmath}
\usepackage{amssymb}
\usepackage{mathtools}
\usepackage{amsthm}

\usepackage[capitalize,noabbrev]{cleveref}

\theoremstyle{plain}

\theoremstyle{definition}

\theoremstyle{remark}

\usepackage[textsize=tiny]{todonotes}

\usepackage{amsmath}
\usepackage{xcolor}
\usepackage{multirow}
\usepackage{colortbl}
\usepackage{booktabs}
\usepackage{multicol}
\usepackage{tcolorbox}
\usepackage{xspace}

\usepackage{epigraph}

\usepackage[ruled,linesnumbered,algo2e]{algorithm2e}
\renewcommand{\algorithmicrequire}{\textbf{Input:}}  
\renewcommand{\algorithmicensure}{\textbf{Output:}} 
\SetKwInOut{Input}{Input}\SetKwInOut{Output}{Output}
\SetKwComment{Comment}{$\triangleright$\ }{}
\usepackage{xcolor}

\definecolor{ForestGreen}{RGB}{34,139,34}
\definecolor{myyellow}{RGB}{181, 181, 27}

\newcommand{\modelname}{Mojito\xspace}

\icmltitlerunning{\modelname: Motion Trajectory and Intensity Control for Video Generation}

\begin{document}

\twocolumn[
\icmltitle{\modelname: Motion Trajectory and Intensity Control for Video Generation }




\icmlsetsymbol{equal}{*}

\begin{icmlauthorlist}
\icmlauthor{Xuehai He}{ucsc}
\icmlauthor{Shuohang Wang}{ms}
\icmlauthor{Jianwei Yang}{ms}
\icmlauthor{Xiaoxia Wu}{ms}
\icmlauthor{Yiping Wang}{uw}
\icmlauthor{Kuan Wang}{gatech}
\icmlauthor{Zheng Zhan}{neu}
\icmlauthor{Olatunji Ruwase}{ms}
\icmlauthor{Yelong Shen}{ms}
\icmlauthor{Xin Eric Wang}{ucsc}
\end{icmlauthorlist}

\icmlaffiliation{ucsc}{University of California, Santa Cruz, USA}
\icmlaffiliation{ms}{Microsoft, USA}
\icmlaffiliation{uw}{University of Washington, USA}
\icmlaffiliation{gatech}{Georgia Institute of Technology, USA}
\icmlaffiliation{neu}{Northeastern University, USA}


\vskip 0.3in
]


\printAffiliationsAndNotice{} 

\begin{figure*}[t!]
  \centering
    \includegraphics[width=\textwidth]{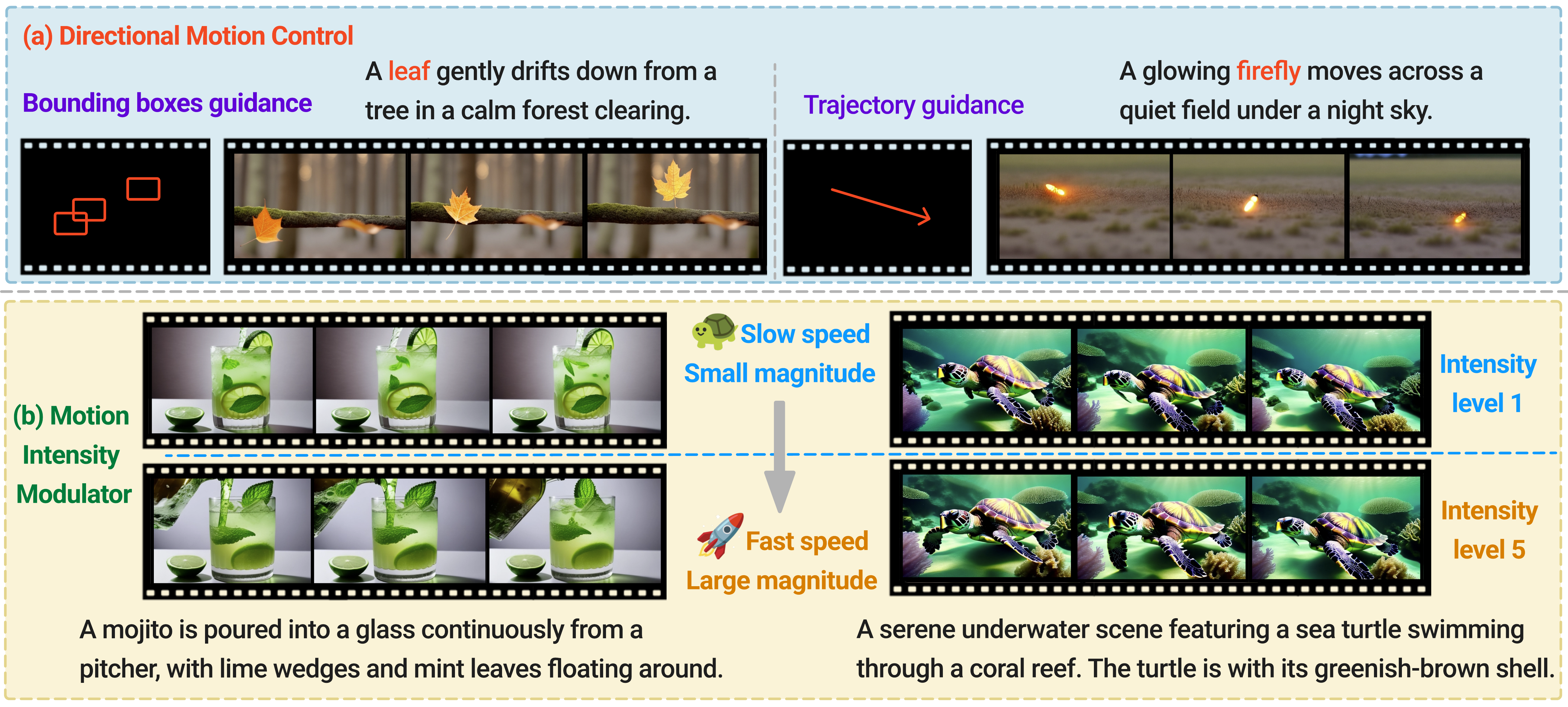}
    \vspace{-4ex}
    \caption{\modelname generates videos that accurately follow specified directions, locations, and trajectories, while adapting to varying input motion intensities. (a) \textbf{\textit{Directional Motion Control}:} the object (\textcolor{red}{leaf, firefly}) in the generated videos can follow input bounding boxes or trajectories over time. (b) \textbf{\textit{Motion Intensity Modulator}:} increasing input motion intensity levels results in a corresponding increase in motion, transforming a relatively static scene into one with more dynamic movement. \textit{Additional examples can be found at~\href{https://sites.google.com/view/mojito-video}{https://sites.google.com/view/mojito-video}.}} 
    \label{Fig:teaser}
\end{figure*}

\begin{abstract}
Recent advancements in diffusion models have shown great promise in producing high-quality video content. However, efficiently training video diffusion models capable of integrating directional guidance and controllable motion intensity remains a challenging and under-explored area. To tackle these challenges, this paper introduces~\modelname, a diffusion model that incorporates both \textbf{mo}tion tra\textbf{j}ectory and \textbf{i}ntensi\textbf{t}y contr\textbf{o}l for text to video generation. Specifically,~\modelname features a~\textit{Directional Motion Control} (DMC) module that leverages cross-attention to efficiently direct the generated object's motion \textbf{without training}, alongside a \textit{Motion Intensity Modulator} (MIM) that uses optical flow maps generated from videos to guide varying levels of motion intensity. Extensive experiments demonstrate~\modelname's effectiveness in achieving precise trajectory and intensity control with high computational efficiency, generating motion patterns that closely match specified directions and intensities, providing realistic dynamics that align well with natural motion in real-world scenarios.  Project page:~\href{https://sites.google.com/view/mojito-video}{https://sites.google.com/view/mojito-video}.
\end{abstract}

\section{Introduction}
\label{sec:intro}
\epigraph{\emph{``Movement is life; control brings mastery.''}}{---Adapted from Aristotle's philosophy}

\begin{figure*}[tp]
  \centering
    \includegraphics[width=0.8\textwidth]{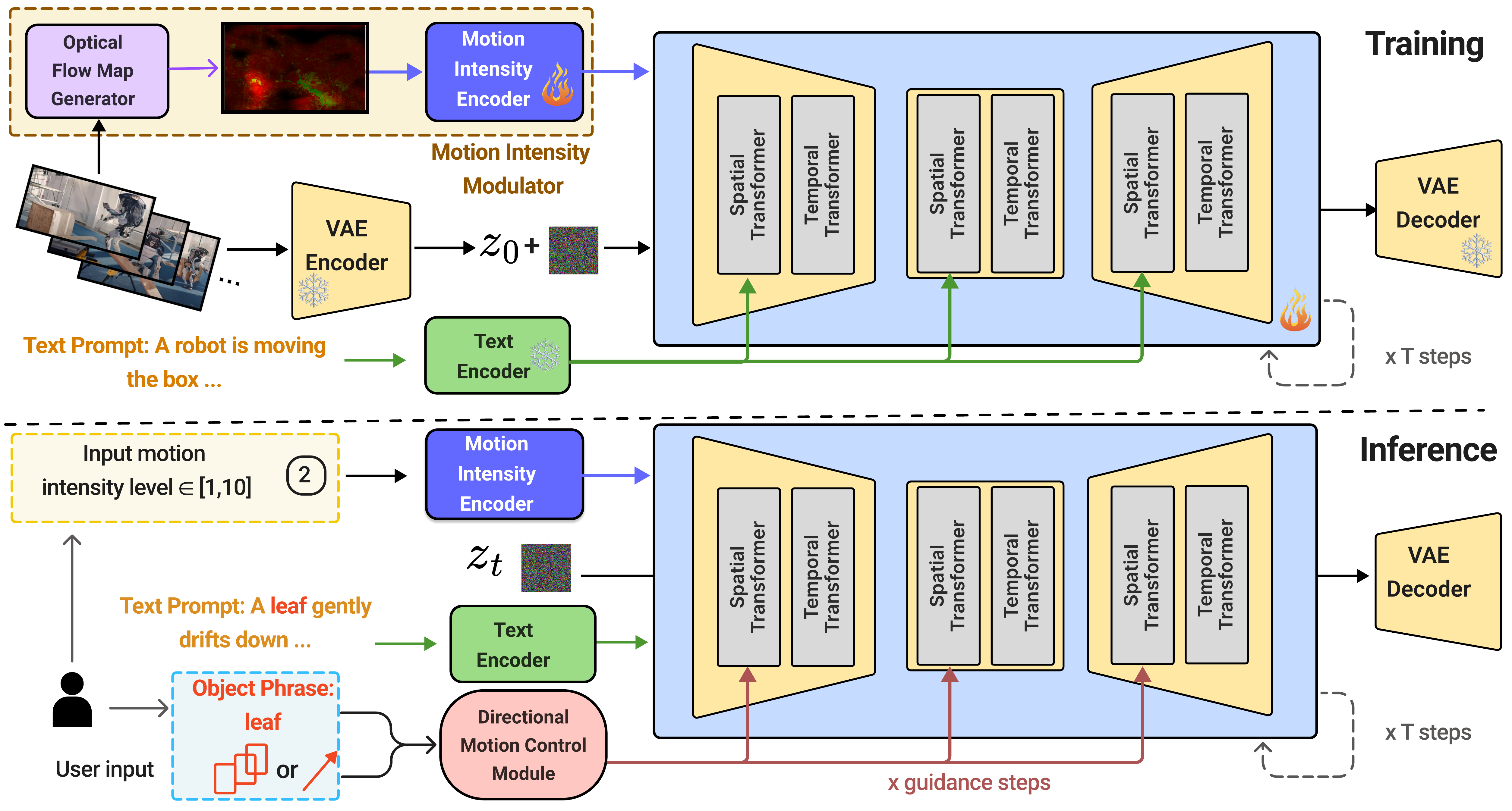}
    \caption{Overview of the \modelname framework. In the training pipeline (top), ~\modelname uses a VAE Encoder to transform input frames into latent features, processed by Spatial and Temporal Transformers within the U-Net. Motion intensity control is introduced through the \textit{Motion Intensity Modulator}, consisting of the Optical Flow Map Generator and the Motion Intensity Encoder. The~\textit{Directional Motion Control} module interprets object phrases within the prompt to align attention with specified trajectories. During inference (bottom), \modelname generates videos following user-defined motion intensity and directional guidance. }
    \label{Fig:overview}
\end{figure*}

Text-to-video (T2V) generation~\cite{stable_video_diffusion,latentvideodiffusion,opensora,opensoraplan,easyanimate2024xu,yang2024cogvideox,vchitect2024} aims to produce diverse, high-quality videos with given text prompts. Unlike image generation~\cite{stable_diffusion,saharia2022photorealistic,nichol2021glide,ramesh2021zero,yu2022scaling,chang2023muse}, which produces a single static frame, video generation can extend beyond visual synthesis and has the potential to serve as a world modeling tool~\cite{nvidia2025cosmos,xiang2024pandora,wang2024world,kang2024far} and a physical AI engine~\cite{Genesis} for simulating real-world dynamics. As world modeling expects video generation models to control and generate the motion of physical objects, enabling realistic interactions with the environment, existing T2V models—despite their creative and powerful generative capabilities—struggle to produce large motion magnitudes~\cite{opensora,opensoraplan,pika2024,modelscope2023wang} or adaptable motion intensities (i.e., motion speed and magnitude) that align with user intent~\cite{nvidia2025cosmos,chen2024videocrafter2}, limiting their applicability.

Towards this, in this work, we aim to address a new research question:~\textit{Can text-to-video models be trained efficiently to control both motion direction and intensity in alignment with user intent?} The primary limitations of current models stem from several key challenges. \underline{First}, existing diffusion-based T2V models—whether U-Net or DiT-based~\cite{dit}—rely on diffusion processes in either pixel space or latent visual feature space during training, but lack dedicated motion modeling mechanisms. Consequently, they inherently struggle to adjust object motion in generated videos solely through text prompts and often fail to maintain temporal consistency in moving objects, resulting in flickering, trajectory inconsistencies, and visual artifacts across frames; \underline{Second}, capturing relative motion data is inherently complex in real-world videos due to the simultaneous movements of both cameras and objects, leading to lack of video data in training. There is also a lack of large-scale, annotated datasets specifically for motion direction and intensity. Existing video datasets rarely include detailed labels for motion dynamics, and obtaining human annotations is labor-intensive; \underline{Third}, training T2V models with such detailed annotations would also bring substantial computational resources.

To tackle these challenges, we present \modelname, the first diffusion model for text-to-video generation that simultaneously integrates text-based prompts with flexible motion controls, allowing for the precise modulation of both motion direction and intensity. Figure~\ref{Fig:teaser} showcases how \modelname enables precise control over both motion trajectory and intensity.
To achieve this, \modelname introduces two novel modules apart of our trained video generation backbone: (1) the \textit{Motion Intensity Modulator} (MIM), which encodes input motion intensities as features and integrates them seamlessly into the diffusion framework during training.  (2) the \textit{Directional Motion Control} (DMC) module, which adjusts object motion direction through cross-attention guidance \textbf{without training}, aligning object trajectories with specified paths.

Our contributions are summarized as follows:
\begin{itemize}
  \item We propose~\modelname, a novel diffusion-based transformer framework for text-to-video generation, integrating both trajectory/direction-aware motion direction control and motion intensity modulation within a unified architecture.
\item We introduce a training-free, test-time~\textit{Directional Motion Control} (DMC) method that grants text-to-video generation with user-specified motion trajectories or directions. Our study explores how to effectively guide attention during video generation by analyzing attention maps and identifying the optimal strategy for steering motion using user-defined inputs, such as bounding boxes, during the forward diffusion process. Furthermore, we show that DMC is a model-agnostic approach that generalizes across different video generation architectures, including U-Net-based and DiT-based~\cite{dit} models.
\item To tackle object inconsistency in generated videos under directional motion control, we introduce a novel temporal smoothness function, optimizing it as it updates latent features during the forward diffusion process to enhance coherence across frames.
\item We develop a training-based~\textit{Motion Intensity Modulator} (MIM) that seeks to achieve precise control over motion intensity by leveraging optical flow~\cite{optical_flow}, ensuring generated videos align with user-defined motion intensity levels. Additionally, we explore three alternative designs for MIM. We find that combining motion intensity embeddings with text embeddings yields the best performance
\item We demonstrate that~\modelname not only generates high-quality videos comparable to state-of-the-art models but also provides effective and efficient motion control. Through ablations, we provide insights into designing efficient, motion-controllable video generation techniques, shedding light on advancing future motion-augmented video generative models.
\end{itemize}
\section{Related Works}
\begin{figure*}[tb]
  \centering
\includegraphics[width=0.95\textwidth]{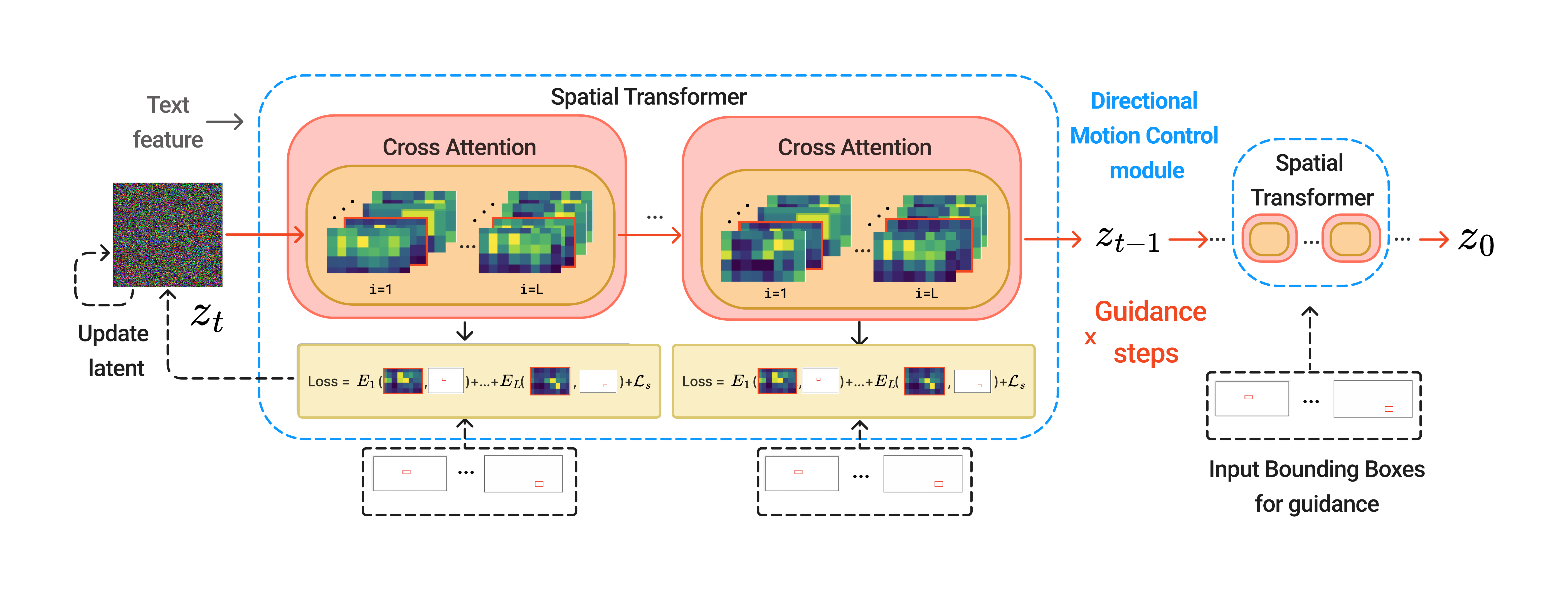}
    \caption{Overview of the~\textit{Directional Motion Control} module. The cross-attention map for the chosen word token in the given guidance step is marked with a red border. We compute the loss and perform backpropagation during inference time to update latents. }
    \label{Fig:dmc}
\end{figure*}

\subsection{Diffusion-based Text to Video Generation}
With the rapid development of generative models, particularly diffusion models, numerous breakthroughs have been achieved in fields such as image generation~\cite{stable_diffusion} and video generation~\cite{latentvideodiffusion,easyanimate2024xu,opensora,opensoraplan,chen2024videocrafter2,yang2024cogvideox,vchitect2024}.  Regarding text-to-video (T2V) models, Make-A-Video~\cite{singer2022makeavideo}, Imagen Video~\cite{imagen2022} are cascaded models, while most
other works, such as LVDM~\cite{latentvideodiffusion}, ModelScope~\cite{wang2023modelscope}, and Align your Latents~\cite{align_latent}, are Stable Diffusion-based
models. They extend the SD framework to videos by incorporating temporal layers to ensure temporal consistency among frames and the spatial parameters are inherited from the pretrained SD UNet. ~\modelname follows a similar architectural approach, using spatial transformers for spatial information and accepting directional control, and incorporating temporal transformers to handle temporal coherence. 

\subsection{Controllable Text-to-Visual Generation}

Recent advancements in text-to-visual generation have enabled more precise and stable control over generated outputs. In image generation, several models~\citep{t2iadapter, composer, unicontrolnet, unicontrol} have been developed to enhance control by leveraging semantic guidance from text prompts alongside various structural conditions. Some approaches utilize attention maps and bounding boxes to manage image layouts, enabling generation based on regional specifications~\cite{Chefer2023AttendandExciteAS, yang2022reco, Li2023GLIGENOG, chen2023trainingfree,training-free-guidance}. ~\citet{mo2024freecontrol} presents a training-free method for controllable image generation; In video generation, which involves additional complexity due to motion, recent works have explored methods for motion control. Training-based methods, such as those in~\citet{zhang2024tora, motionctrl}, allow for trajectory-based motion control, while~\citet{jain2024peekaboo} achieves video control through designed masking in the diffusion process. Unlike these approaches, our method leverages attention maps in a training-free manner for directional motion control. Our model can control the movement of specific objects within the generated video, and uniquely, it enables motion intensity control—providing a new level of flexibility in video generation not addressed in previous works.

\section{Preliminaries}

\paragraph{Latent Video Diffusion.} 
Latent Diffusion Models (LDMs), such as Stable Diffusion~\cite{stable_diffusion}, consist of two main components: an autoencoder and a denoising network. The autoencoder encodes images into a latent space, while the denoising network operates within this latent space, progressively transforming random noise into a coherent image representation. Building on this, Latent Video Diffusion Models (LVDMs)~\cite{latentvideodiffusion} extend LDMs to video generation by incorporating temporal dynamics in the latent space. They are trained using a noise-prediction objective, minimizing the difference between predicted and actual noise added to the data:
\begin{equation}
    \mathcal{L}_{LDM} = \mathbb{E}_{i} \left[ \| \epsilon - \epsilon_\theta(z_t, t, c) \|_2^2\right],
\end{equation}
where \( \epsilon_\theta(\cdot) \) is the noise prediction function of the U-Net, and the condition \( c \) is provided to the U-Net through cross-attention, enabling control based on input text or other conditioning signals. The latent state \( z_t \) follows a Markov process, with Gaussian noise progressively added at each step to the initial latent state \( z_0 \):
\begin{equation}
    z_t = \sqrt{\bar{\alpha}_t} z_0 + \sqrt{1 - \bar{\alpha}_t} \epsilon, \quad \epsilon \sim \mathcal{N}(0, I),
\end{equation}
where \( \bar{\alpha}_t = \prod_{i=1}^t (1 - \beta_i) \) and $\beta_t$ is a step-dependent coefficient that controls noise strength at time $t$.

\paragraph{Motions in Video Generation.}\quad 
Motion in videos generally consists of two main components: \textit{direction} and \textit{intensity}. The direction of motion is derived from the trajectory or bounding boxes that an object follows across frames, while the intensity (or strength) reflects the speed and amplitude of movement along this path.

We define a motion trajectory as a sequence of spatial positions:
$
\mathcal{T} = \left\{(x_{1}, y_{1}), (x_{2}, y_{2}), \ldots, (x_{L}, y_{L})\right\},
$
where each $(x_{i}, y_{i})$ represents the object’s position in frame $i$, with $x \in [0, W)$ and $y \in [0, H)$, where $L$ is the total number of frames, and $W$ and $H$ are the width and height of the video frame, respectively.


\section{The~\modelname Video Generation Framework}
\label{sec:motion_guided_video_generation}
\subsection{Methodology Overview}
Figure~\ref{Fig:overview} illustrates the architecture of~\modelname, which comprises two core modules for motion control: the \textit{Directional Motion Control} (DMC) module and the \textit{Motion Intensity Modulator} (MIM). These two modules work together to allow controlled motion in terms of direction and intensity, as well as to coordinate interaction with the generated video content. We first introduce the test-time, training-free ~\textit{Directional Motion Control}. Then we will introduce the~\textit{Motion Intensity Modulator} and finally the~\modelname backbone.

\subsection{Training-Free Directional Motion Control}
Existing text-to-video generation models often struggle to achieve user-defined motion direction, such as following a specific trajectory for an object. We address this limitation by introducing a method that dynamically guides motion direction during inference-time sampling, generating samples from a conditional distribution $p(z | y, \mathcal{T}, i)$ with additional directional controls. Here, $z$ represents the generated latent video, $y$ is the input text, and $\mathcal{T}$ specifies a user-defined input trajectory associated with certain tokens $y_n$. This directional control is achieved by modifying attention responses within cross-attention layers in the spatial transformer block of~\modelname, as shown in Figure~\ref{Fig:dmc}.

Cross-attention layers are essential for managing spatial layouts in generated content~\cite{prompt_to_prompt, training-free-guidance, training-layout-control}. The attention score $A_{u,n}^{(\gamma)}$ between spatial location $u$, text token $y_n$ in cross-attention layer $\gamma$ determines their association, with $\sum_{n=1}^N A_{u,n}^{(\gamma)} = 1$. This competitive interaction among text tokens helps guide object motion by biasing attention maps. By aligning attention focus along the trajectory $\mathcal{T}$ with the token $y_n$, we gain fine-grained frame-by-frame control over object placement without requiring additional training.

Specifically, for a given input trajectory $\mathcal{T} = \{(x_{i}, y_{i})\}_{i=1}^{L}$, where $L$ is the number of points in the trajectory, we expand each point $(x_{i}, y_{i})$ to create a bounding box $B_{i}$ for frame $i$. This bounding box is defined as:
\begin{equation}
B_{i} = \{ (x, y) : |x - x_{i}| \leq \Delta_x, |y - y_{i}| \leq \Delta_y \},
\end{equation}
where $\Delta_x$ and $\Delta_y$ are tolerance values that define the size of the bounding box around each trajectory point. These bounding boxes $B_{i}$ are allocated for each video frame $i$, ensuring alignment with the trajectory.

To align the attention maps with the bounding boxes over the course of the video, we define a frame-specific energy function $E_{i}$ at each video timestep $i$:
\begin{equation}
\label{energy_function}
E_{i}\left(A_{i}^{(\gamma)}, B_{i}, n\right) = \left(1 - \frac{\sum_{u \in B_{i}} A_{i,u,n}^{(\gamma)}}{\sum_u A_{i,u,n}^{(\gamma)}}\right)^2,
\end{equation}
where $A_{i,u,n}^{(\gamma)}$ denotes the attention score at video timestep $i$ in layer $\gamma$, spatial location $u$, and text token $y_n$ for the input object phrase. This function encourages attention to concentrate within the bounding box $B_{i}$ at each timestep $i$, thereby achieving effective alignment of the attention with the trajectory throughout the video sequence.

\paragraph{Temporal Smoothness Function.}\quad 
To ensure temporal coherence and avoid abrupt changes in motion, we introduce a temporal smoothness function across frames, defined as the expectation of squared differences in attention maps over consecutive frames:
\begin{equation}
\label{temporal_loss}
\mathcal{T}_s = \mathbb{E}_{i} \left[ \| A_{i,:,:}^{(\gamma)} - A_{i-1,:,:}^{(\gamma)} \|^2 \right],
\end{equation}
where $A_{i,:,:}^{(\gamma)}$ and $A_{i-1,:,:}^{(\gamma)}$ are attention maps for consecutive frames $i$ and $i-1$. Optimizing this function penalizes large frame-to-frame attention changes, promoting smooth motion across frames.

\paragraph{Gradient Update.}\quad 
The latent variable $\boldsymbol{z}^{(t)}$ at diffusion timestep $t$ is iteratively updated using the gradients of the combined energy functions:
\begin{equation}
\label{gradient_update}
\boldsymbol{z}^{(t)} \leftarrow \boldsymbol{z}^{(t)} - \sigma_t^2 \eta \nabla_{\boldsymbol{z}^{(t)}} \sum_{i=1}^{L} \sum_{\gamma \in \Gamma} \left( E_{i}\left(A_{i}^{(\gamma)}, B_{i}, n\right) + \lambda \mathcal{T}_s \right),
\end{equation}
where $\eta > 0$ controls the guidance strength, $\lambda$ controls the scale of the temporal smoothness function, and $\sigma_t = \sqrt{\frac{1 - \alpha_t}{\alpha_t}}$ adjusts for the noise level at diffusion timestep $t$. Scaling by $\sigma_t^2$ ensures that early diffusion timesteps with high noise receive weaker guidance, which intensifies as noise decreases in later timesteps, refining latent representations near the end of the diffusion process.

To generate a video, we alternate between gradient updates and denoising steps, iteratively refining the latent variable $\boldsymbol{z}^{(t)}$ to align cross-attention maps with the intended trajectory. The combination of iterative guidance and smoothness enforcement allows the model to produce coherent video sequences with precise spatial control. The overall algorithm framework of the DMC module is showcased in Algo.~\ref{alg:motion_guided_video_generation}.

\subsection{Motion Intensity Modulator}

The~\textit{Motion Intensity Modulator} (MIM) enables controlled adjustment of motion intensity in generated videos. During training, we compute the optical flow~\cite{optical_flow} strength between consecutive frames to quantify motion intensity, using it as a conditioning input for the diffusion model. This conditioning allows the model to capture varying levels of motion intensity, guiding the generation process based on the specified motion dynamics.

\paragraph{Optical Flow Map Generator.}\quad 
To quantify motion intensity in videos, we compute optical flow maps that capture the temporal dynamics between consecutive frames. These maps encode pixel-wise motion direction and magnitude, providing a structured representation of movement. For each video, frames are processed sequentially and converted to grayscale to facilitate efficient computation using the Farneback method~\cite{farneback2003two}. This approach estimates motion by analyzing pixel displacements between adjacent frames. The resulting magnitude values serve as a measure of motion intensity, which we normalize and discretize into integer levels ranging from $1$ to $10$. 

\paragraph{Motion Intensity Encoder.}\quad 
To incorporate the motion intensity into our model, we convert the optical flow strength of videos into embeddings. Similar to~\citet{kandala2024pix2gif}, rather than feeding the motion intensity as a raw numerical value, we transform it into a word form during training. This aligns with the CLIP model’s preference for text-based representations, as CLIP serves as our text encoder.

The motion intensity is embedded through a learned embedding layer \( \mathcal{M} \), which maps the motion intensity to an embedding vector \( c_M \). This vector is duplicated and concatenated with itself to ensure that its influence is strong enough when combined with the text embeddings.
\paragraph{Training and Loss Function.}\quad 
The model is trained by minimizing the following loss function:
\begin{equation}
\mathcal{L}_{LDM}^{\prime} = \mathbb{E}\left[\left\|\epsilon - \epsilon_\theta\left(z_t, t, c_T, c_M\right)\right\|_2^2\right].   
\end{equation} Here, \( c_M \) represents the motion intensity condition and \( c_T \) the text condition, which together provide a comprehensive conditioning signal for guiding the generation of temporally consistent, motion-controlled videos.

\paragraph{Inference.}\quad 
During inference, for motion intensity guidance, \modelname takes two inputs: a text prompt, and the motion intensity. These inputs are processed in two pathways: one directly through the diffusion model and the other through the MIM module. The text embedding $c_T$ is obtained by processing the text prompt via the CLIP text encoder, while the motion embedding $c_M$ is generated through the motion intensity embedding layer. The final conditioning input $c_L = c_T + c_M$ combines the text and motion embeddings, allowing the generation process to be guided with both semantic and motion-based conditioning.

\subsection{Model Backbone}
The backbone of~\modelname is based on a Conv-Spatial-Temporal Transformer architecture. The model structure consists of input blocks, a mid-block, and output blocks, each containing convolutional layers, temporal transformers, and spatial transformers. This design enables handling motion control in both the temporal and spatial dimensions. The details of model backbone are given in the~\textsc{Appendix}.

The~\textit{Temporal Transformer} processes video data along the time axis, capturing dependencies between frames by reshaping the input into temporal sequences. This block includes relative position encoding and optional causal masking for unidirectional attention, making it effective for capturing motion patterns across frames; The~\textit{Spatial Transformer} captures spatial relationships within each frame. Using linear or convolutional projections, this block reshapes and processes spatial information to learn localized patterns, which is essential for high-resolution video generation. Cross-attention layers enable alignment with external context when available, further refining spatial coherence in the generated content. The~\textit{Directional Motion Control} (DMC) module operates within Spatial Transformer to enforce trajectory-based motion guidance.

\section{Experiments}
\subsection{Datasets}
We use Panda70M~\cite{panda70m}, InternVID~\cite{internvid}, and Mira~\cite{ju2024miradata} to train both the~\modelname backbone and the Motion Intensity Modulator module. From Panda-70M, we extract a high-quality 10M subset for training. Additionally, we process 1,000 videos from Panda-70M for evaluation of motion intensity modulation. For motion direction guidance evaluation, we utilize the LaSOT dataset~\cite{fan2021lasot}, which provides paired text descriptions, bounding box annotations, and corresponding video sequences. For InternVID and Mira, we process and use the entire dataset for training. Further details on dataset preprocessing and usage can be found in the~\textsc{Appendix}.

\begin{figure}[tb]
  \centering
\includegraphics[width=\linewidth]{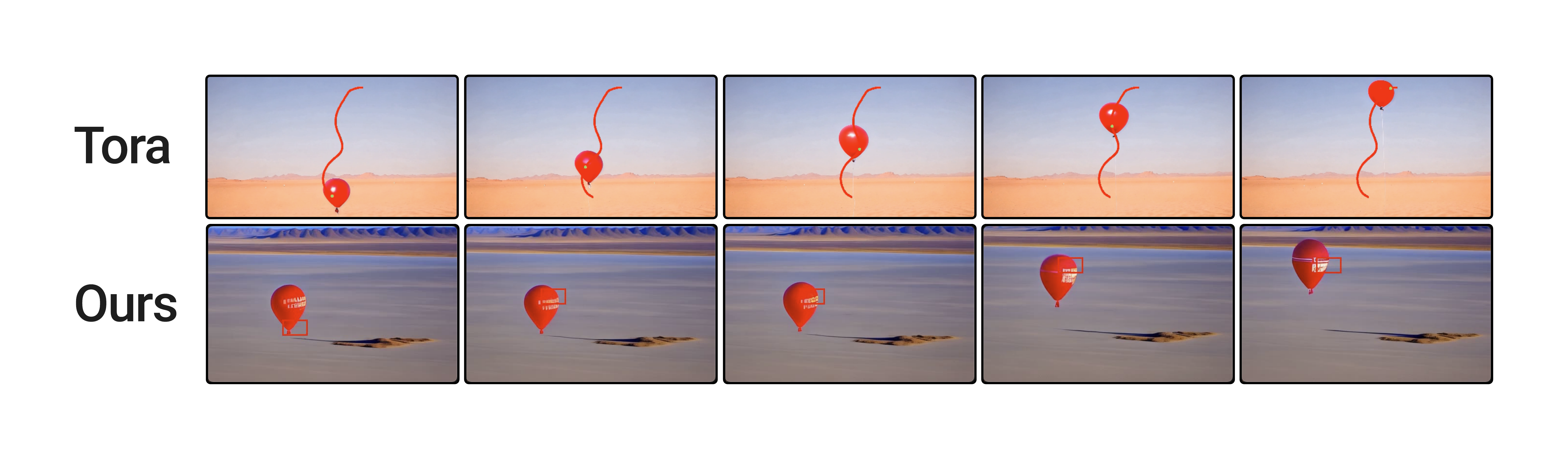}
\vspace{-4ex}
   \caption{Qualitative comparison of directional control with Tora. ~\modelname achieves motion control comparable to Tora while offering additional capabilities to specify objects and precise locations \textbf{without training}. The red bounding boxes, serving as inputs to~\modelname, guide the balloon to follow the specified trajectory.}
    \label{Fig:comparison}
\end{figure}

\subsection{Evaluation Metrics}
For quantitative evaluation, we employ commonly used metrics: Fréchet Video Distance (FVD)~\cite{unterthiner2019fvd} and CLIPSIM~\cite{dynamicrafter}, which calculates the average similarity across all generated frames in relation to the input caption. Specifically, we compute the FVD for a set of 300 videos, each consisting of 16 frames. 

To evaluate the effectiveness of directional guidance, we utilize the off-the-shelf OWL-ViT-large open-vocabulary object detector~\cite{oject_detector} to extract bounding box annotations from the generated videos. We then compute the Intersection-over-Union (IoU)~\cite{iou} between the detected and target bounding boxes to measure spatial alignment. We also compute the Centroid Distance (CD) as the distance between the centroid of generated objects and input bboxes, normalized to 1.
For motion intensity control, we compute the difference between the detected average optical flow in the generated video and the target motion intensity.

\subsection{Implementation Details}
\noindent \textbf{Model Backbone.}\quad  For the Variational Autoencoder (VAE) framework, the encoder processes input frames at a resolution of 320$\times$512, progressively downsampling through a series of residual blocks and convolutional layers with channel multipliers of $(1, 2, 4, 4)$. This results in a latent representation with a spatial resolution of $40\times64$. We use CLIP (ViT-H-14)~\cite{clip} as the text encoder, allowing for high-quality text-to-video alignment. For video processing, spatial and temporal downsampling factors of 8 are applied in both dimensions. The attention matrices (such as $q$ and $kv$) within the spatial transformer are structured as $(t, b, c)$, where $t$ represents the temporal length. Trajectory adjustments are applied only within the mid-block and the initial layers of the upsampling branch in the denoising U-Net~\cite{unet}, which provides a balance between maintaining video quality and achieving precise motion control. Additional implementation details and parameter settings are provided in the \textsc{Appendix}.

\noindent \textbf{Training.}\quad We use the DeepSpeed~\cite{deepspeed} framework to train our model, utilizing memory and computation optimizations for distributed training. The model training is carried out using the FusedLamb~\cite{fusedlamb} optimizer combined with the LambdaLRScheduler. We use a batch size of 4 and a base learning rate of $1.0 \times 10^{-5}$. More details can be found in the \textsc{Appendix}.

\subsection{Performance Evaluation}

\begin{table*}[ht]
\centering
\caption{Human evaluation between \modelname, OpenSora~\cite{opensora}, OpenSora plan~\cite{opensoraplan}, and VideoCrafter2~\cite{chen2024videocrafter2}. It includes 2400 comparisons on 400 video pairs. Each video pair contains 5 ratings from different human annotators on Amazon Turk. The detailed setup is given in the \textsc{Supplementary Materials}.}
\label{HumanEvaluation}
\scalebox{0.85}{
\begin{tabular}{@{}l|ccc|ccc|ccc@{}}
\toprule 
& \modelname & Tie & OpenSora   & \modelname & Tie & VideoCrafter2   & \modelname & Tie & OpenSora plan   \\
\midrule
\midrule
Direction Alignment &  \textbf{84.8\%} & 3.3\% & 12.0\% & \textbf{88.0\%} & 5.0\% & 7.0\% & \textbf{86.8\%} & 3.0\% & 10.3\% \\
Intensity Alignment & \textbf{46.8\%} & 22.3\% & 31.0\% & \textbf{40.0\%} & 28.5\% & 31.5\% & \textbf{43.5\%} & 23.8\% & 32.8\% \\
\bottomrule
\end{tabular}}
\end{table*}

\begin{figure}[tbp]
  \centering
\includegraphics[width=\linewidth]{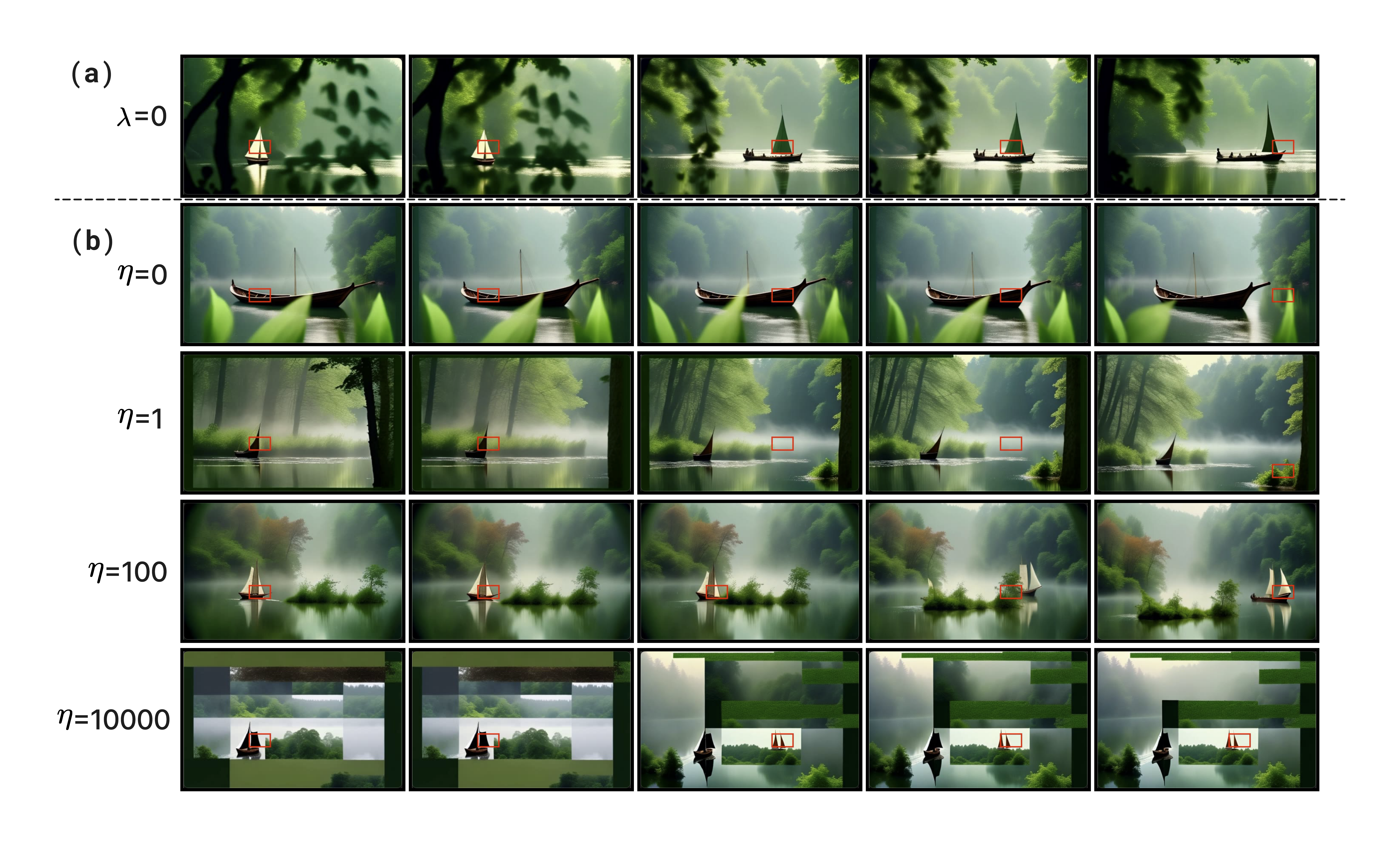}
\vspace{-5ex}
    \caption{ (a) \textbf{Ablation Study on Temporal Smoothness Loss}: Without temporal smoothness loss, the generated sailboat exhibits inconsistencies across frames. (b) \textbf{Ablation Study on Guidance Strength}: Adjusting the guidance strength demonstrates a trade-off between video quality and trajectory alignment.}
    \label{Fig:direction}
\end{figure}

\begin{figure}[tbp]
  \centering
\includegraphics[width=\linewidth]{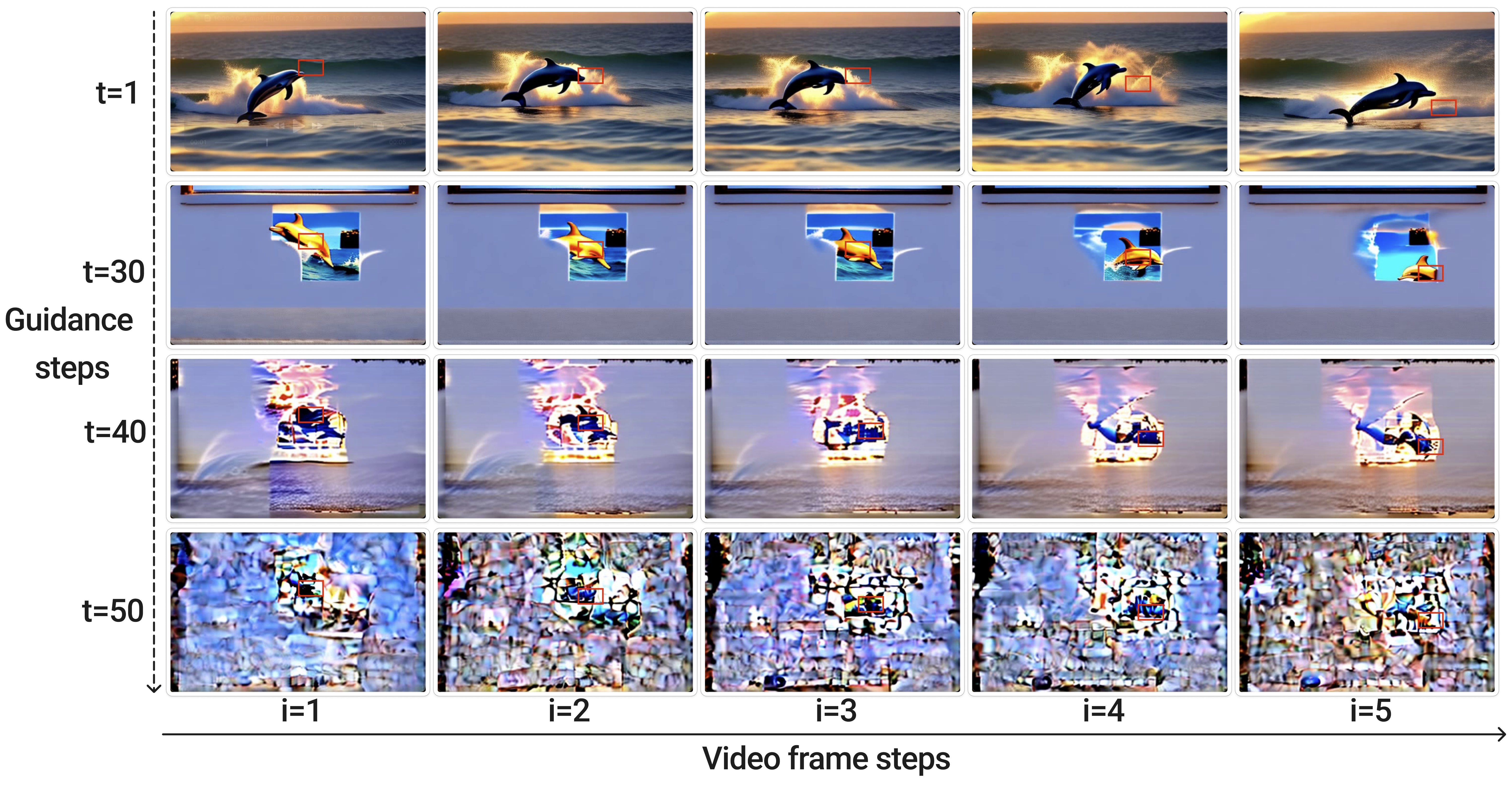}
\vspace{-4ex}
    \caption{ Ablation study on the effect of guidance steps. While increasing the number of guidance steps improves trajectory alignment, excessive guidance steps can degrade overall video quality.}
    \label{Fig:guidance_steps}
\end{figure}

\begin{table}[tbp]
\centering
\caption{Quantitative evaluation of trajectory alignment and video quality with different total numbers of guidance steps. Setting higher guidance steps generally enhances trajectory alignment while maintaining video quality.}
\label{tab:ablation_direction}
\resizebox{0.47\textwidth}{!}{
\begin{tabular}{lccccc}
\toprule
\multirow{2}{*}{Method} & \multirow{2}{*}{FVD $\downarrow$} & \multirow{2}{*}{CLIPSIM $\uparrow$} & \multicolumn{3}{c}{Direction} \\
\cmidrule(lr){4-6}
 & & & mIoU $\uparrow$ & AP50 $\uparrow$ & CD $\downarrow$ \\
\midrule
\midrule
$t=1$ & 423 & 0.2520 & 18.0 & 11.2 & 0.34  \\
$t=5$ & \textbf{422} & 0.2511 & 24.5 & 14.4 & 0.32  \\
\rowcolor[gray]{0.9} 
$t=10$ & \textbf{422} & \textbf{0.3700} & 26.0 & 17.1 & \textbf{0.28} \\
$t=30$ & 479 & 0.1880 & \textbf{28.3} & \textbf{18.9} & \textbf{0.28} \\
\bottomrule
\end{tabular}}
\end{table}

\begin{table}[tbp]
\centering
\caption{Quantitative evaluation of different designs for MIM. Combining motion intensity embeddings with text embeddings achieves the best overall performance, balancing video quality, semantic similarity, and motion alignment. Treating motion intensity embeddings as a global conditional input shows good motion alignment but significantly degrades video quality.}
\label{tab:ablation_intensity}
\resizebox{\linewidth}{!}{
\begin{tabular}{l|ccc}
\toprule
Method & FVD $\downarrow$ & CLIPSIM $\uparrow$ & Motion Alignment $\downarrow$ \\
\midrule
\midrule
w/o MIM & 421 & 0.250 & 0.307 \\
Global Conditional Input & 438 & 0.250 & 0.097 \\ 
Direct Text Input & \textbf{422} & 0.256 & 0.174\\
\midrule
\rowcolor[gray]{0.9}Combined with Text Embedding& 423 & \textbf{0.273} & \textbf{0.089} \\
\bottomrule
\end{tabular}}
\end{table}

\noindent \textbf{Qualitative Evaluations on Directional Control.}
We present a qualitative comparison with Tora~\cite{zhang2024tora} in Figure~\ref{Fig:comparison}. The text prompt used is ``{A \textcolor{red}{red helium balloon} floats slowly upward into the sky over a barren desert and expansive Gobi landscape.}" The benefits of \modelname are threefold. \underline{First}, \modelname features a \textit{training-free capability}, enabling precise directional control at test time without requiring additional training data or fine-tuning. Unlike training-based approaches like Tora, \modelname can dynamically adjust object trajectories during sampling at inference time, ensuring robust control without added computational cost in training.
\underline{Second}, \modelname provides \textit{fine-grained directional control} by allowing users to specify motion paths for named objects. This level of control contrasts with baseline models like Tora, which lack the capability to follow object-specific directional prompts and typically follow arbitrary trajectories. In Figure~\ref{Fig:comparison}, we showcase scenarios where \modelname is given specific object names (e.g., ``red helium balloon") and precise locations, enabling it to guide each object’s motion along user-defined paths. \underline{Third}, the DMC module in \modelname is model-agnostic and easily pluggable into different architectures, including DiT-based models, with the results shown in Figure~\ref{Fig:cosmos}.

\noindent \textbf{Human Evaluation.}\quad
We conduct human evaluations to assess the effectiveness of~\modelname in both motion intensity control and directional control. For evaluating the Motion Intensity Modulator, we compare~\modelname with baselines by rewriting the text prompts for baselines to explicitly specify the desired intensity level. The results of this evaluation are presented in Table~\ref{HumanEvaluation}. Details on the evaluation and settings are provided in the \textsc{Appendix}.

\subsection{Visualization of Attention Maps}
To understand the progression of directional control in \modelname, we examine cross-attention maps at different sampling steps within the denoising network. The text prompt is ``A small red ball bounces along a cobblestone path lined with greenery." As shown in Figure~\ref{Fig:attention}, the attention maps for the target object (the ``red ball") initially display random patterns at the beginning of the sampling process. However, as sampling steps progress, the attention gradually refines and starts to align with the bounding boxes defined for each video frame step. This refinement ensures that the generated video closely follows the user-defined trajectory.

\subsection{Ablation Study}

\subsubsection{Designs in Directional Motion Control}
In this section, we analyze the effects of various configurations for directional control, specifically examining different values of guidance strength ($\eta$), temporal smoothness function ($\lambda$), and guidance steps. The results are presented in Figure~\ref{Fig:direction}. The text prompt used is ``{A vintage wooden {sailboat} glides steadily down a mist-covered river.}" and ``A {dolphin} leaps through the waves of the ocean at sunrise."

\noindent \textbf{Temporal Smoothness Function.}\quad 
When $\lambda = 0$ (no temporal smoothness function), the generated video exhibits noticeable changes in views and inconsistent sailboat appearances. 

\noindent \textbf{Guidance Strength.}\quad 
With $\eta = 0$ (no guidance strength), the generated ``sailboat" does not follow specified bounding boxes at all. As $\eta$ increases, the sailboat begins to follow bounding boxes more closely. The best results reaches with $\eta = 100$, where the sailboat’s trajectory aligns well with the bounding boxes while maintaining good visual quality. However, setting $\eta$ to an excessively high value (e.g., $\eta = 10000$) results in severe degradation of video quality, with visible artifacts such as square patterns.

\noindent \textbf{Guidance Steps.}\quad
We perform an additional experiment to evaluate the effect of varying the number of guidance steps on video quality and trajectory alignment. The results, shown in Table~\ref{tab:ablation_direction}, indicate that increasing the number of guidance steps from $t=1$ to $t=10$ generally enhances trajectory alignment without significantly impacting video quality. However, when the number of guidance steps becomes too large (e.g., $t \geq 30$), the video quality deteriorates substantially. Qualitative results illustrating this effect are provided in Figure~\ref{Fig:guidance_steps}.

\subsubsection{Alternative Choices for Motion Intensity Modulator}

In this section, we explore various designs for the~\textit{Motion Intensity Modulator} (MIM). The primary approach in this paper involves combining a motion intensity embedding with text prompt embeddings and feeding them into~\modelname. Additionally, we experimented with several alternatives:

\noindent \textbf{Without MIM.}\quad Removing MIM during training and testing. In this setup, we perform inference by rewriting the text prompts to include the intensity level. 

\noindent \textbf{Global Conditional Input.}\quad Combining the intensity embedding with FPS and time embeddings, using this global embedding as a conditioning input for convolution layers. 

\noindent \textbf{Direct Text Input with Fine-tuning.}\quad In this approach, we rewrite the text prompts to include the motion intensity level and then fine-tune the model using this new data. Both training and inference are guided by the modified text input that explicitly specifies the motion intensity.

To compare each design, we evaluated the alignment of optical flow between the generated and input videos, as well as video quality and semantic similarity metrics. The quantitative results for each design are shown in Table~\ref{tab:ablation_intensity}, indicate that combining motion intensity embeddings with text embeddings yields the best performance. This demonstrates that treating motion intensity as an additional conditioning input and incorporating it into the cross-attention alongside text embeddings is a highly effective strategy.

\section{Conclusion}
\label{sec:conclusion}

In this work, we presented~\modelname, a novel framework for text-to-video generation that enables integrated control over motion trajectory/direction and intensity. By introducing the~\textit{Directional Motion Control} module and the~\textit{Motion Intensity Modulator}, ~\modelname allows specifying object trajectories and motion intensity during inference. The DMC module is training-free, while the MIM module does not require additional motion data annotations, making it the first approach to explicitly model motion intensity in text-to-video generation and can be trained efficiently. Experimental results demonstrate that \modelname effectively produces high-quality, motion-controlled videos while remaining generalizable and orthogonal to other existing approaches.

\section*{Impact Statement}
In this paper, the video generation framework may be misused to generate misleading or harmful content~\cite{deepfakes} and raises concerns about fairness and potential biases in generated outputs. To mitigate these risks, Mojito can be deployed with safeguards such as watermarking generated content, implementing harmful video filters to ensure appropriate usage, and addressing biases in training data to promote fairness and inclusivity in its applications.

\nocite{langley00}

\bibliography{example_paper}
\bibliographystyle{icml2025}

\newpage
\appendix
\onecolumn
\newpage
\appendix
\onecolumn


\section{Model Architecture}
\begin{figure*}[t]
  \centering
\includegraphics[width=0.85\linewidth]{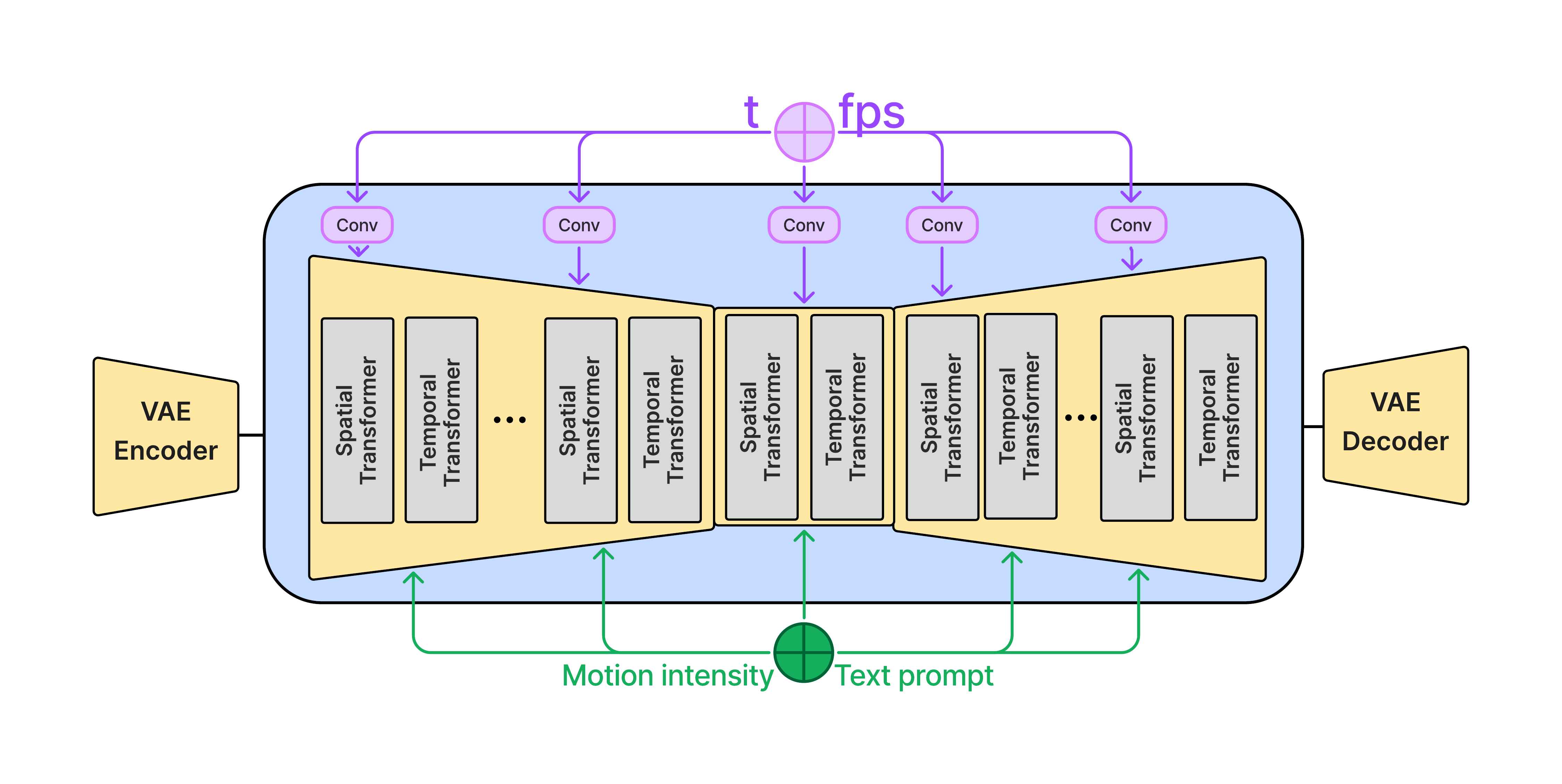}
    \caption{The architecture of~\modelname. The framework combines a VAE Encoder-Decoder structure with spatial and temporal transformers to process spatial and temporal information effectively. Motion intensity embeddings and text prompt embeddings condition the cross-attention layers, providing flexible and controllable motion direction and strength. Temporal convolutional layers handle features like frame per second (fps) and timesteps (t), allowing the model to generate temporally coherent and spatially consistent videos.}
    \label{fig:model_architecture}
\end{figure*}

We illustrate the architecture of~\modelname backbone in Figure~\ref{fig:model_architecture}. The model builds upon a VAE-based latent diffusion framework and incorporates both spatial and temporal transformers within the U-Net architecture to achieve fine-grained control over motion direction and intensity. Spatial transformers process spatial information within each frame, while temporal transformers ensure temporal coherence across video frames. 

Motion intensity embeddings and text prompt embeddings are integrated into the cross-attention layers, enabling the model to conditionally generate videos based on user-specified motion properties. Additional convolutional layers handle the temporal features like frame per second (fps) and  timesteps (t), enhancing the flexibility of the model to adapt to diverse inputs.

\section{Additional Qualitative Results with Cosmos}
\begin{figure*}[t]
  \centering
\includegraphics[width=\textwidth]{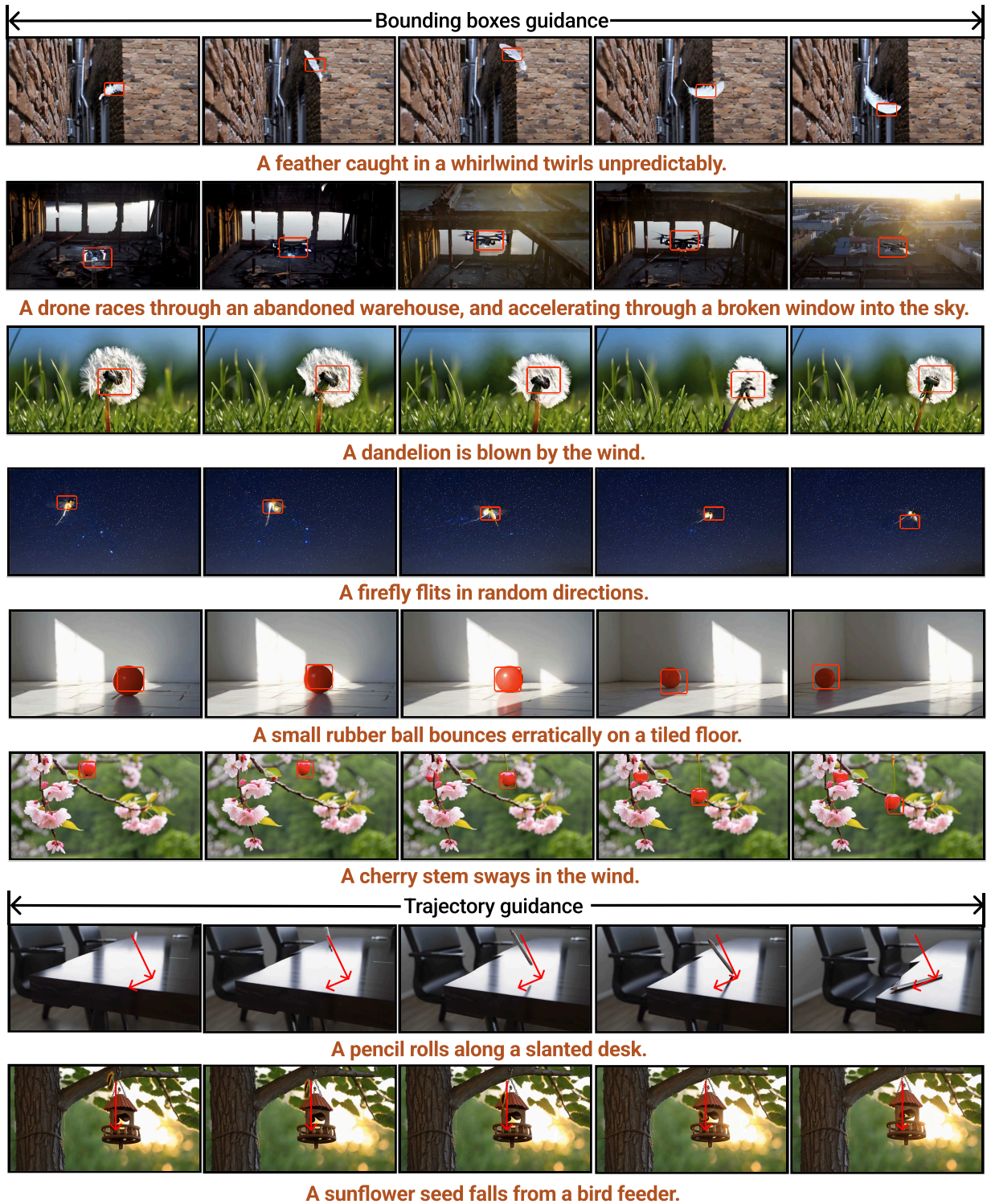}
    \caption{Additional examples demonstrate the application of the DMC module on the DiT-based video generation backbone, Cosmos~\cite{nvidia2025cosmos}. The model successfully follows both bounding box guidance and trajectory constraints
    }
    \label{Fig:cosmos}
\end{figure*}

We provide additional examples generated by applying the DMC module on Cosmos~\cite{nvidia2025cosmos} using the 7B diffusion version in Figure~\ref{Fig:cosmos}. The objects in the generated videos strictly follow the input bounding boxes and trajectory guidance, indicating the DMC module is model-agnostic and easy to plug in.

\section{Visualizations on Attention Map}
\begin{figure*}[t]
  \centering
\includegraphics[width=\textwidth]{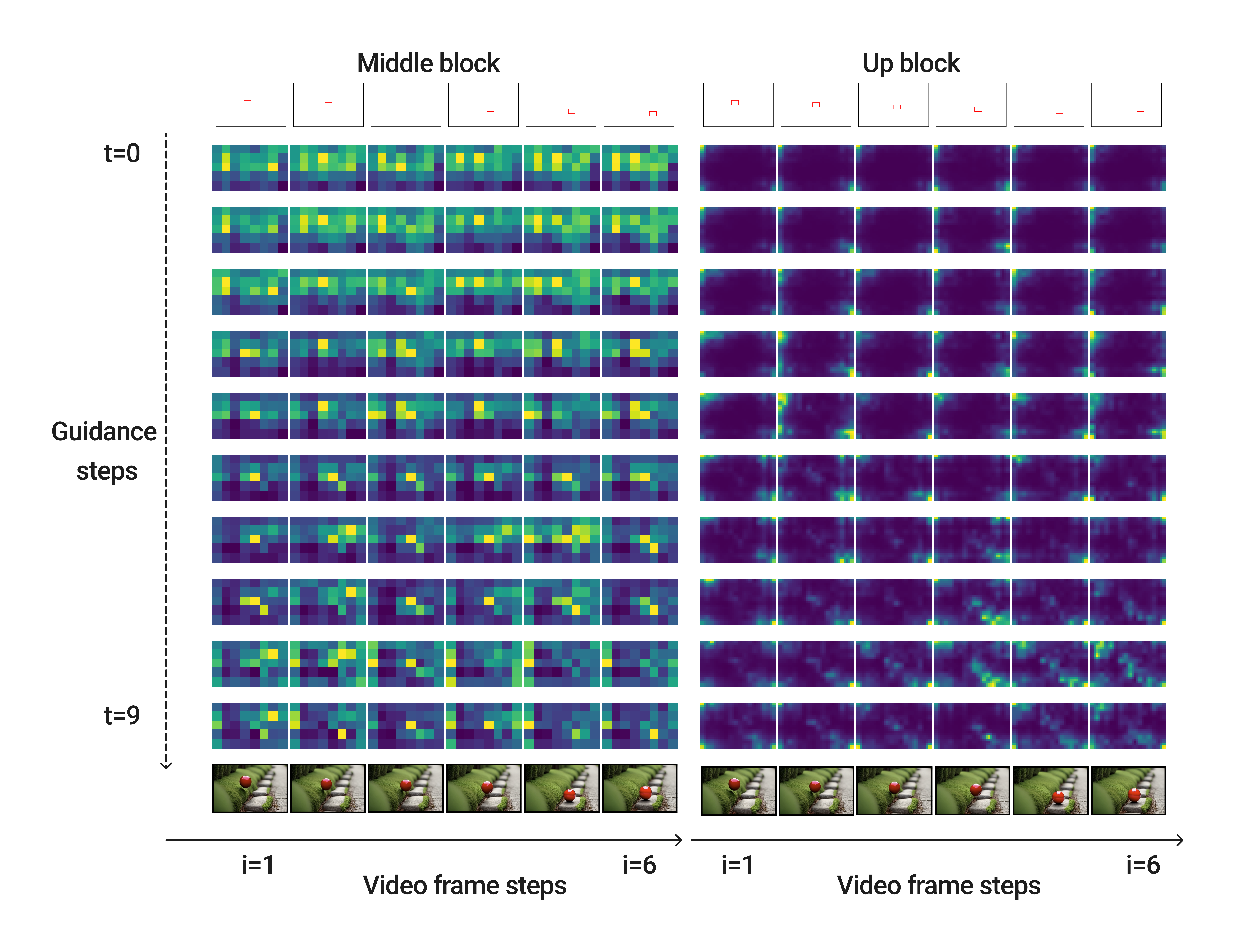}
    \caption{Progression of cross-attention maps during sampling steps for directional guidance. Initially, the attention map for the red ball shows random patterns, but as sampling progresses, it increasingly aligns with the bounding boxes set for each video timestep.
    (Zoom in for better visualization.)
    }
    \label{Fig:attention}
\end{figure*}

We conduct more visualizations on cross-attention maps at different sampling steps within the denoising network in Figure~\ref{Fig:attention}.
The attention maps for the target object (the ``red ball") initially display random patterns at the beginning of the sampling process. However, as sampling steps progress, the attention gradually refines and starts to align with the bounding boxes defined for each video frame step.

\section{ Algorithm Framework}
The detailed algorithm framework for the DMC module is presented in Algo.~\ref{alg:motion_guided_video_generation}.

\section{Dataset}
We use Panda 70M~\cite{panda70m}, InternVID~\cite{internvid}, and Mira~\cite{ju2024miradata} datasets to train~\modelname for motion intensity control. For InternVID and Mira, we follow the official data processing setups.

For Panda 70M, we process the data in two stages to ensure semantic consistency and quality. In the first stage, we split long videos using PySceneDetect with a content detector and a cut-scene threshold of 25 frames, introducing artificial cuts every 5 seconds for long, uncut footage. To filter clips with semantic transitions, we compute ImageBind~\cite{girdhar2023imagebind} features at 10\% and 90\% of each clip and retain only those with a Euclidean distance $\leq 1.0$. 

In the second stage, consecutive clips with similar semantic content are merged by comparing their boundary features and concatenating those with a feature distance $\leq 0.6$. Additional post-processing ensures clip quality and diversity: clips shorter than 2 seconds or containing minimal motion are excluded, and clips longer than 60 seconds are trimmed. We further filter clips to ensure semantic diversity (feature distance $> 0.3$) and remove unstable segments by trimming the first and last 10\% of each clip. This results in 70,817,169 clips with an average duration of 8.5 seconds.

\section{Implementation Details}
\subsection{Training Details}
For the \modelname backbone, we utilize the pretrained VAE encoder from~\cite{chen2024videocrafter2}. The training process integrates spatial and temporal transformers alongside the MIM module to ensure spatial consistency and controlled motion across video frames. 

The model is trained for a maximum of 5,000,000 steps with a batch size of 1 and gradient accumulation over 2 steps. Logging occurs every 50 steps, and checkpoints are saved every 1,000 steps, with additional periodic saving every 5,000 steps to evaluate and pick the best model checkpoint. 

We utilize PyTorch Lightning for distributed training with DeepSpeed Stage-2 optimization, using 32-bit precision for stability. Memory optimizations include overlapping communication and computation, ensuring efficient usage of GPU resources. We adopt a fixed learning rate of $1 \times 10^{-3}$, optimized using a step-based LambdaLR scheduler. The scheduler applies a linear decay from the initial step, reducing the learning rate to 1\% of its original value over 20,000 steps. Gradient clipping is enforced with a maximum norm of 0.5 to stabilize training. Mixed precision training is employed for computational efficiency, with DeepSpeed stage-2 optimization enabled for effective memory management. The model is trained on a large-scale cluster using 16 nodes with a total of 128 A100 GPUs.

\subsection{Model Architecture Configuration}
 The autoencoder processes input frames at a resolution of $320 \times 512$, downsampling them into a latent space with a spatial resolution of $40 \times 64$ and a channel depth of 4. The conditional input is encoded using a FrozenOpenCLIPEmbedder~\cite{openclip}, which remains fixed during training to leverage text prompts effectively for generation. The model also integrates additional conditioning inputs, such as FPS information, to improve motion control.

\begin{algorithm}[!t]
\caption{Training-Free Directional Motion Control}
\label{alg:motion_guided_video_generation}
\Input{Input text $y$, object phrase $y_n$, user-defined trajectory $\mathcal{T}$, guidance strength $\eta$, temporal smoothness weight $\lambda$, final guidance timestep $t_g$
}
\Output{Generated video from $\boldsymbol{z}^{(0)}$}

Initialize latent variable $\boldsymbol{z}^{(T)} \sim \mathcal{N}(0, I)$
\For{$t \leftarrow T$ \KwTo $t_g$}{
    Compute noise level $\sigma_t = \sqrt{\frac{1 - \alpha_t}{\alpha_t}}$ 

    Compute attention scores: $A_{i}^{(\gamma)} = \text{Attention}(y_n, u, i)$ 

    Define trajectory-aligned region: $B_{i} = \text{Region}(\mathcal{T}, i)$ 

    Compute energy function: 
    $
    E_{i}\left(A_{i}^{(\gamma)}, B_{i}, n\right) = \left(1 - \frac{\sum_{u \in B_{i}} A_{i,u,n}^{(\gamma)}}{\sum_u A_{i,u,n}^{(\gamma)}}\right)^2
    $
    ~\Comment*[r]{\textcolor{blue}{Eq.~\ref{energy_function}}}

    Compute temporal smoothness loss: 
    $
    \mathcal{T}_s = \mathbb{E}_{i} \left[ \| A_{i,:,:}^{(\gamma)} - A_{i-1,:,:}^{(\gamma)} \|^2 \right]
    $
    ~\Comment*[r]{\textcolor{blue}{Eq.~\ref{temporal_loss}}}

    Compute total gradient:
    $
    G = \nabla_{\boldsymbol{z}^{(t)}}\sum_{i=1}^{L} \sum_{\gamma \in \Gamma} \left( E_{i}\left(A_{i}^{(\gamma)}, B_{i}, n\right) + \lambda \mathcal{T}_s \right)
    $

    Update latent variable:
    $
    \boldsymbol{z}^{(t)} \leftarrow \boldsymbol{z}^{(t)} - \sigma_t^2 \eta G
    $
}

\tcc{\textcolor{blue}{Perform directional guidance steps until $t_g$}}
\Return Generated video from $\boldsymbol{z}^{(0)}$
\end{algorithm}

\begin{figure*}[tb]
\centering \includegraphics[width=\textwidth]{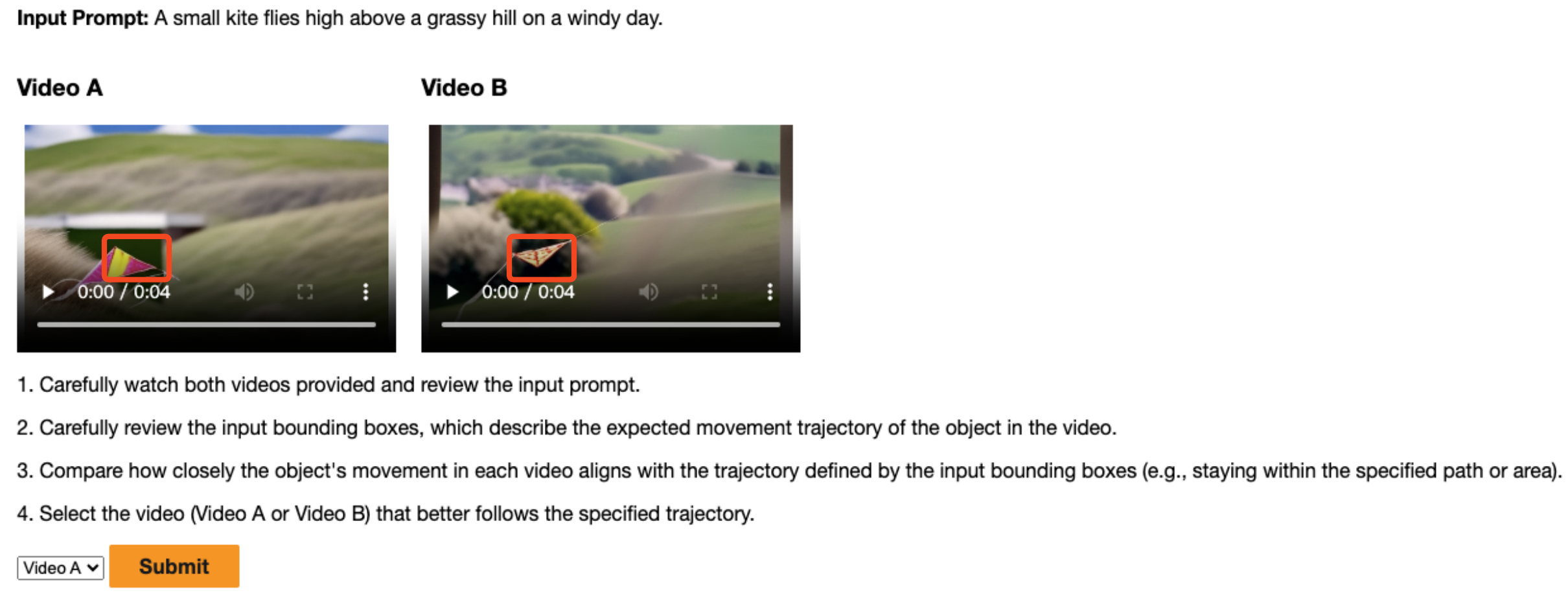} \caption{Human evaluation interface for assessing motion direction using Amazon Turkers.} \label{fig:direction_eval
} 
\end{figure*}

\section{Human Evaluation Interface}

In Figure~\ref{fig:direction_eval
}, we present the human evaluation interface used on Amazon Mechanical Turk for assessing motion intensity. Participants are asked to rate the perceived motion intensity in the generated videos on a predefined scale.

\begin{figure*}[tb] \centering \includegraphics[width=\textwidth]{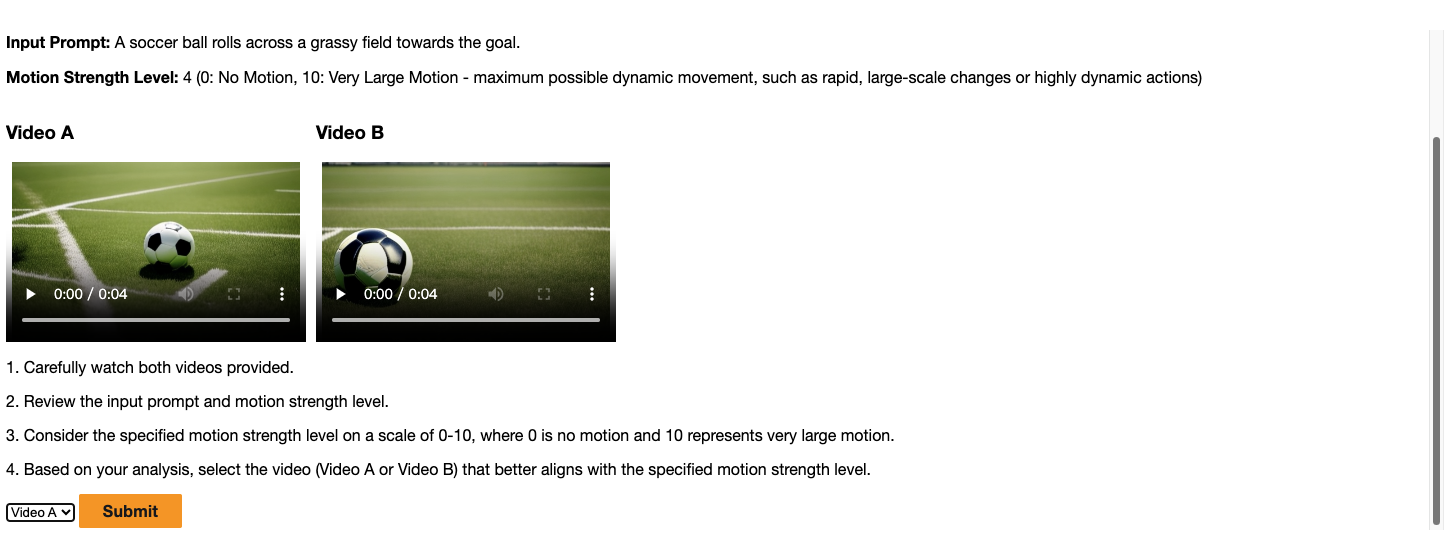} \caption{Human evaluation interface for assessing motion intensity using Amazon Turkers.} \label{fig:intensity_eval
} \end{figure*}

In Figure~\ref{fig:intensity_eval
}, we display the human evaluation interface used for evaluating motion direction alignment. Participants are instructed to assess how well the motion direction in the videos aligns with the specified trajectories.

\begin{figure*}[tb]
    \centering
    \includegraphics[width=\textwidth]{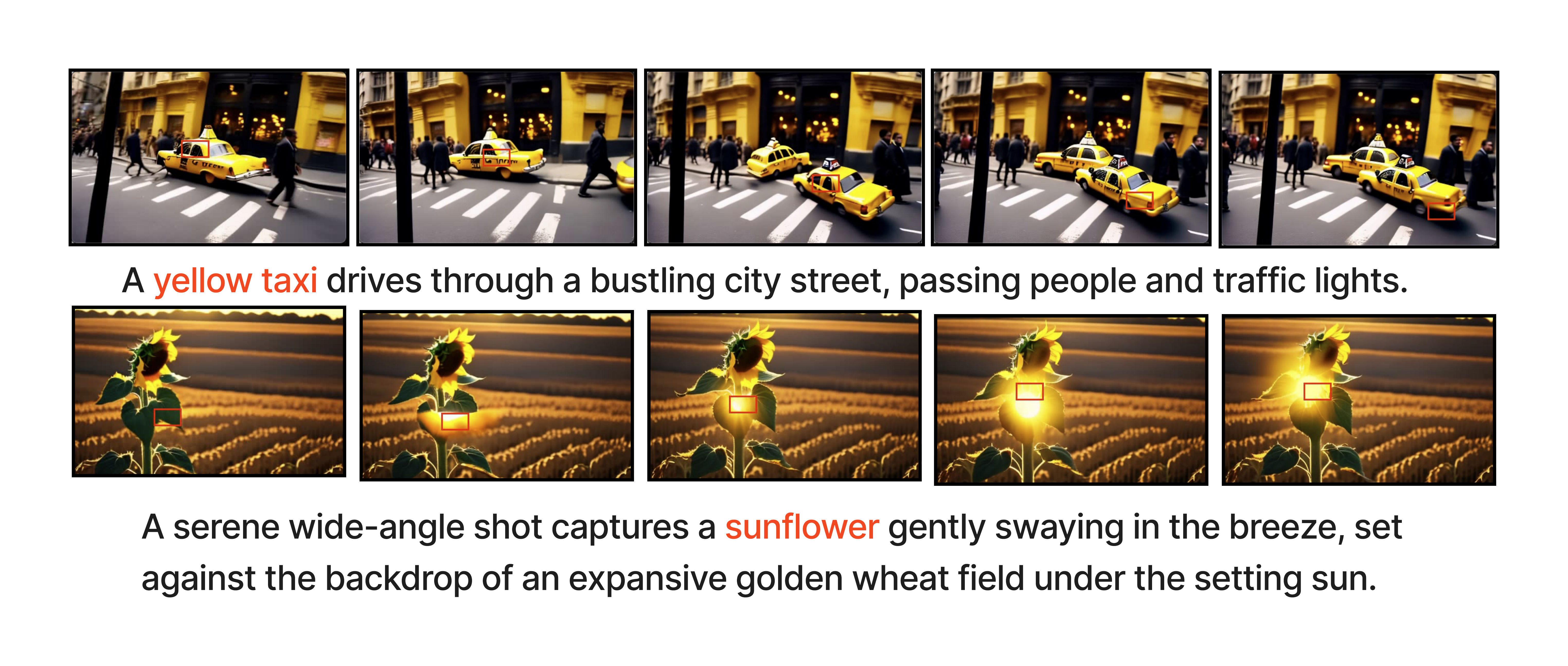}
    \caption{Failure cases for directional motion control. The first case occurs when the bounding boxes move significantly faster across frames than the video frame rate. 
The second failure case arises when the text prompt contains similar objects, such as a ``\textcolor{red}{sunflower}" and a ``sun." In these instances, the guidance mechanism may mistakenly switch focus to the wrong object, resulting in unintended transitions,
such as directing attention from the sunflower to the sun.}
    \label{failure}
\end{figure*}

\section{Failure Cases}
Figure~\ref{failure} illustrates two common failure cases observed in directional motion control. The first case occurs when the bounding boxes move significantly faster than the video frame rate. In such situations, multiple objects may appear simultaneously, causing confusion. For example, instead of the first ``\textcolor{red}{yellow taxi}" moving into the bounding box as intended, a second yellow taxi may appear at the later bounding box location. 

The second failure case arises when the text prompt contains similar objects, such as a ``\textcolor{red}{sunflower}" and a ``sun." In these instances, the guidance mechanism may mistakenly switch focus to the wrong object, resulting in unintended transitions, such as directing attention from the sunflower to the sun.


\end{document}